# Hand Pointing Detection Using Live Histogram Template of Forehead Skin

Ghassem Tofighi[1], Nasser Ali Afarin[2], Kamraan Raahemifar[3], Anastasios N. Venetsanopoulos[4]

[1,3,4] Dept. of Electrical and Computer Engineering
Ryerson University
Toronto, Canada

[2] Dept. of Computer Engineering
The University of Isfahan
Isfahan, Iran

E-mail: gtofighi@ryerson.ca, afarin@live.com, kraahemi@ee.ryerson.ca, tasvenet@ryerson.ca

*Abstract*— Hand pointing detection has multiple applications in many fields such as virtual reality and control devices in smart homes. In this paper, we proposed a novel approach to detect pointing vector in 2D space of a room. After background subtraction, face and forehead is detected. In the second step, forehead skin H-S plane histograms in HSV space is calculated. By using these histogram templates of user's skin, and back projection method, skin areas are detected. The contours of hand are extracted using Freeman chain code algorithm. Next step is finding fingertips. Points in hand contour which are candidates for the fingertip can be found in convex defects of convex hull and contour. We introduced a novel method for finding the fingertip based on the special points on the contour and their relationships. Our approach detects hand-pointing vectors in live video from a common webcam with 94%TP and 85%TN.

*Keywords: Hand Posture Recognition, Hand Pointing Recognition, Finger Orientation Detection, Back projection.*

## I. INTRODUCTION

In many applications of Human Computer Interaction (HCI), we desire to send commands to the computer and receive results. There are a lot of physical devices such as mouse and keyboard for this purpose. However, it is more convenient to use some natural approaches to communicate with computers. Two main approaches are speech recognition and body and hand gesture recognition methods for communicating with computers. One of the most important hand gestures with variety of applications is hand pointing gesture. Hand pointing gesture and its direction can be used in Virtual Reality (VR), smart homes, household robot and any application which needs pointing gesture. Another use of pointing gesture is for helping in speech recognition applications when we want to specify the location parameter of a verbal sentence.

In this paper, we introduce a novel method to detect pointing gesture and its direction in live sequence of images from a common webcam in a room. Our system is designed to detect hand pointing in natural home environment and real time. In this approach, we use the following main tasks to detect pointing direction:

- Background subtraction
- Face detection skin histogram template
- Hand detection
- Contour processing for pointing gesture detection

### 1-1- Related work

There are extensive research on body features and hand gestures. In [1], Tofighi et al. uses histogram template of skin for detecting hand in the image based on H-S plane histograms. In [2], they have extended their research on finding 10 different hand postures using HandReader dataset. In [3], Wren et al. demonstrate the system Pfinder which employs a statistical model of color and shape to obtain a 2D representation of head and hands. Azarbayejani and Pentland [4] describe a 3D head and hands tracking system. This system calibrates automatically from watching a moving person. Pointing detection also had done in [7] for first time. In [7], Mehmet Gokturk et all introduce a mounted device to detect pointing direction. Massaki Fukumoto et al in [6], use image processing to detect pointing detection. More than that, Hand pointing recognition is addressed in [6] by introducing a finger-controlled joystick for using as a mouse. Jojic et al. [7] detect and estimate pointing gestures in dense disparity maps. In [8], a method for recognizing hand orientation in a non-consistent environment has introduced, which employs user eye orientation and tracking finger for calculating the orientation of hand. Afarin et all [9], introduce an algorithm to detect hand pointing using adaptive histograms.

## II. OUR APPROACH

In many approaches of on hand gesture recognition, there are 4 main steps as follows: (1) Skin detection (2) Hand detection (3) Feature extraction (4) Feature processing for gesture detection. We are using the same procedure.

Our approach is developed for using in natural environment of home. In this situation, there are a variety of colors and shapes which are similar to different parts of the human body.

Our proposed approach has 5 steps to achieve the goal. These steps are (1) Background subtraction (2) Face detection (3) Skin detection (4) Hand pointing gesture detection (5) Hand pointing direction detection. These steps will be described in the following sections.

### 2-1- Background subtraction

For background subtraction, we should have an appropriate model from background to segment foreground from the background; In fact, the heart of all background subtraction algorithms is making a model of background. These algorithms are divided into two categories: Non-recursive Techniques and Recursive Techniques. In our research, we are using recursive approach. There are also multiple versions of recursive techniques. Two main techniques are Gaussian filter and CodeBook filter.

*Gaussian filter*: We provide a brief description of the popular scheme used in [12]. The internal state of the system is described by the background intensity $B_t$ and its temporal derivative $B'_t$, which are recursively updated as follows in equation (1):

(1) $$\begin{bmatrix} B_t \\ B'_t \end{bmatrix} = A.\begin{bmatrix} B_{t-1} \\ B'_{t-1} \end{bmatrix} + K_t.I_t - H.A.\begin{bmatrix} B_{t-1} \\ B'_{t-1} \end{bmatrix}$$

Matrix A describes the background dynamics and H is the measurement matrix. Their particular values used in [26] are as follows in equation (2):

(2) $$A = \begin{bmatrix} 1 & 0.7 \\ 0 & 0.7 \end{bmatrix}, H = \begin{bmatrix} 1 & 0 \end{bmatrix}$$

The Kalman gain matrix $K_t$ switches between a slow adaptation rate $a_1$ and a fast adaptation rate $a_2 > a_1$ based on whether $I_{t-1}$ is a foreground pixel in eqation (3):

(3) $$\begin{cases} K_t = \begin{bmatrix} a_1 \\ a_1 \end{bmatrix} & \text{if } I_{t-1} \text{ is foreground} \\ K_t = \begin{bmatrix} a_1 \\ a_1 \end{bmatrix} & \text{otherwise} \end{cases}$$

*CodeBook* : Sample background values at each pixel are quantized into codebooks which represent a compressed form of background model for a long image sequence. This allows to capture structural background variation due to periodic-like motion over a long period of time under limited memory. The method described in [13] can handle scenes containing moving backgrounds or illumination variations.

The CB algorithm adopts a quantization/clustering technique [14], to construct a background model. Samples at each pixel are clustered into the set of codewords. The background is encoded on a pixel by pixel basis.

The key features of algorithm described in [13] are in the followings: (1) resistance to artifacts of acquisition, digitization and compression, (2) capability of coping with illumination changes, (3) adaptive and compressed background models that can capture structural background motion over a long period of time under limited memory, (4) unconstrained training that allows moving foreground objects in the scene during the initial training period.

Simple algorithm for Background Subtraction using codebook described in [12] is:

1) x = (R,G,B), I ← R + G + B
2) For all codewords in M in Eq.1 find the codeword $c_m$ matching to x based on two conditions:
    (a) colordist(x; $v_m$) ≤β
    (b) brightness(I,<$Ĭ_m, Î_m$>) = true
3) $BGS(x) = \begin{cases} foreground & if\ there\ is\ no\ match \\ background & otherwise \end{cases}$

Where

- BGS(x) is subtraction operation for an incoming pixel x.
- The two conditions (a) and (b), detailed in [13], are satisfied when the pure colors of $x_t$ and $c_m$ are close enough and the brightness of $x_t$ lies between the acceptable brightness bounds of $c_m$. Instead of finding the nearest neighbour, we just find the first codeword to satisfy these two conditions. β is the sampling threshold (bandwidth).

We implemented all approaches described in this section and result shows that CodeBook algorithm performs better. Extensive comparison between codebook and some other subtraction algorithm done in [13]. They also announced Codebook performs better. An example of running this algorithm is depicted in figure 1.

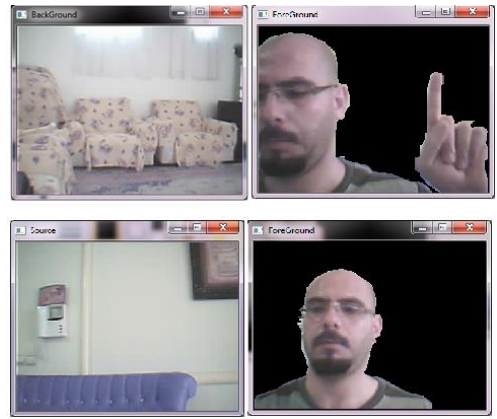

Figure 1. Two example of background subtraction using codebook algorithm

We use frame differencing to track light changes. If 80% of pixel values in grayscale image of the scene change α units, and α is less than 10, we update the model. If α is more than 10 units, we first remove 90% of model data and afterwards update the model with new frames. For updating the model, we update model pixels that were not in the last frame and marked as foreground. The experiments shows our model converges after 30 to 45 frames.

### 2-1- Forehead detection

Forehead detection is performed to find a template for user's skin. This template is used for the histograms which is necessary for the next step. To achieve this goal, we first need to detect the face. Viola-Jones detector [15] is used for this purpose. Although there are multiple face detection algorithms such as [16], [17], [18], [19] with more accuracy and lower speed, Viola-Jones face detector [15] is preferred because it is fast and accurate enough for our purpose. Viola-Jones has a technical report [20] which shows it can also detect multi view faces very fast.

After detecting the face we find forehead as follows. Suppose the face detector gives us a rectangle with
- p(x,y) which is top left corner
- size(width, height) as size of rectangle.

By viewing the face detector results we saw that forehead for most of person is located in a rectangle with
- p2(x+width/3,height/10) as top left corner
- size(width/3,height/5) as size.

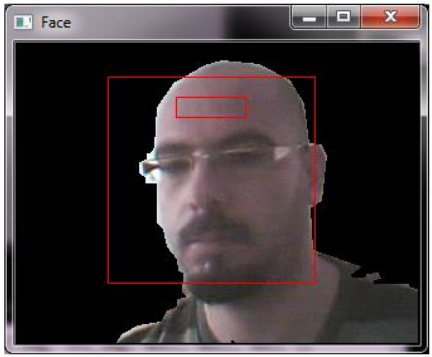

Figure 2. Forehead detection

Figure 2 shows forehead detection. Our experiments show that Viola-Jones algorithm performs well with illumination and light colour changes. So we can have a skin model which is almost robust in light illumination change.

### A. Skin detection

Although in background subtraction step we eliminate background pixels, there are still some other noise pixels. For example, in a foreground image which user enters the camera view, we have his/her clothes, and we should segment hand from this foreground image.

There are lots of skin detection algorithms which use difference color spaces for skin area. For instance, [21] is one of the thresholding algorithms that use HSV space. In [22] they use $YC_RC_B$ space and [23] uses neural network for skin detection using four different colour spaces: RGB, Normalized RGB, $YC_RC_B$ and YUV and concluded Normalized RGB is better than RGB for skin detection.

Some skin detection algorithms described in [24], [25] and [26], which is a method for skin detection using a proposed skin histogram and a pre-learning step. Tofighi et all [1], described a method for hand detection that we used it to detect skin areas more accurate and robust to light change. We use normalized RGB of input image for training images. In this algorithm, hand segmentation has been carried out using back projection method [26]. It performs well in real time. We use forehead skin area detected in previous section as skin area template. Histogram of these patches of skin areas are calculated. By using these histograms, for each pixel, we can calculate the probability of the belonging to skin or non-skin categories as follows:

(4) $$P(rgb|skin) = \frac{s[rgb]}{T_s}$$

and

(5) $$P(rgb|!skin) = \frac{n[rgb]}{T_n}$$

Where in equations 8 and 9 $s[rgb]$ is $rgb$ pixels in $H_s$ and $n[rgb]$ is rgb pixels in $H_n$, and $T_s$ and $T_n$ are the number of all pixels in $H_s$ and $H_n$ respectively. Therefore, we can propose Equation 10 for the probability of being skin pixel given its color.

(6) $$P(skin|rgb) = \frac{p(rgb|skin)*p(skin)}{p(rgb|skin)*p(skin)+ p(rgb|!skin)*p(!skin)}$$

where $P(skin) = \frac{T_s}{T_s+T_n}$ and $P(!skin)= 1- p(skin)$.

The Back Projection algorithm calculates the back project of the histogram. That is, similarly to calculate histogram algorithm, at each location (x, y) the algorithm collects the values from the selected channels in the input images and finds the corresponding histogram bin. However, instead of incrementing it, the function reads the bin value. We can divide each value by summation of histogram values. In this way, we have probability of each pixel. This value is in [0...1] domain. We can show histogram values by scale it to [0...255]. In terms of statistics, the algorithm computes probability of each element value in respect with the empirical probability distribution represented by the histogram. Figure 3 shows skin area probability calculated by Back Projection.

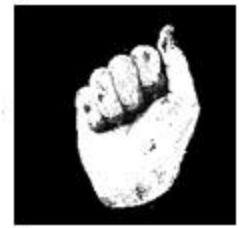

Figure 3. Back Projection of skins using forehead histogram [1]

As shown in figure 5, the brighter areas are more probable to be skin area (as they actually are), whereas the darker areas have less probability. Please note that these "dark" areas belong to surfaces that have some shadows.

Next step is thresholidng the back projected image to suppress weak colors. It may also have sense to suppress pixels with non-sufficient color saturation and too dark. We can use binary thresholding filter with threshold θ for back projected image to remove areas with low probability.

Finally, we obtain the blob representation of the hand by applying a connected components algorithm to the image, which groups pixels into the same blob.

### 2-2- Hand pointing gesture detection

For detecting pointing gesture, we should remove noises and afterwards, find hand area and its contour. Using hand contour features, fingertip will be detected.

For each connected component discovered in previous section, we find its contour with an algorithm such as Freeman chain code in clockwise or counterclockwise order. This algorithm find boundary points of a blob or connected component. After finding the contour of blob, we should calculate the area of the blob. We can use equation 7 for this purpose.

(7) $$A = \frac{1}{2}\sum_{i=0}^{n-1}(x_i y_{i+1} - x_{i+1} y_i)$$

Afterwards, we select three blobs which have more areas. The biggest one is head and the two others are hand. The left one is left hand and right one is right hand.

After hand detection we use two features of the blob to find fingertips Center of gravity of the blob, and Convex hull and convex defects.

*Center of Gravity*

For finding center of gravity of the segmented hand blob, we can use Equation 8 which is used for calculating moments:

(8) $$m_{p,q} = \iint_{-\infty}^{\infty} x^p y^q f(x,y) dx dy \quad p,q = 0,1,2,\ldots$$

where $f(x,y)$ is the value of pixel $(x,y)$. In the current problem we have:

(9) $$f(x,y) = \begin{cases} 1 & skin \\ 0 & non-skin \end{cases}$$

After calculating $m_{00}$, $m_{01}$, and $m_{10}$ we can calculate the center of gravity (COG), $(\bar{x}, \bar{y})$, where $\bar{x} = \frac{m_{10}}{m_{00}}$ and $\bar{y} = \frac{m_{01}}{m_{00}}$.

Figure 4 shows the center of gravity in the segmented hand.

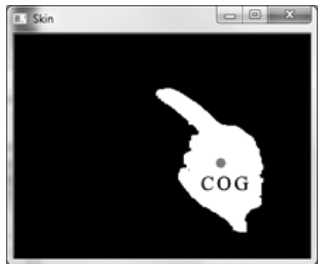

Figure 4. Center of Gravity (COG)

As we mentioned earlier, edge contour is a set of points, which are sorted in counterclockwise order around the center of gravity, $C = \{(x_i, y_i)\}$, $i=1..n$.

For all points $(x_i, y_i)$ in $C$, we calculate the angle created by three points, $(x_{i-k}, y_{i-k})$, $(x_i, y_i)$, and $(x_{i+k}, y_{i+k})$. For calculating these angles, we compute inner product of vectors created by $(x_{i-k}, y_{i-k})$, $(x_i, y_i)$, and $(x_i, y_i)$, $(x_{i+k}, y_{i+k})$. If the angles between these three points is less than a threshold, $\theta_t$, we inference $(x_i, y_i)$ is a corner in hand edge contour and it should be fingertip or a hole between fingers.

In our system, after normalizing all contours by its length, experiments have shown best values for $k$ and $\theta_t$ are 16 and 30 respectively. We insert all these important corners in a list, $L_1$.

For all points in $L_1$, we calculate distance to the COG and if we find a dominant maximum which has distinctly greater than other distances, we inference it should be the fingertip. In this system, we suppose a distance as dominant maximum distance, if it is maximum distance and its difference by the second maximum is greater than 1/6 length of the contour.

*Using convex hull and convex defects*

Convex hull are points of contour that make a closed curve that all contour points are into it. Convex defects are points of contour that has maximum distance to line that through two sequential point of convex.

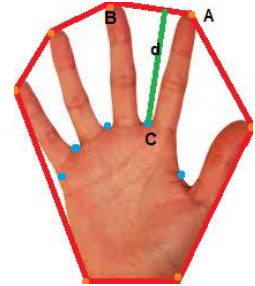

Figure 5. Convex hull and convex defects

In figure 5, yellow points are convex hull points and blue ones are convex defects. In contour points between point A and point B, point C has maximum distance from line that connect A and B.

The distance between $C(x_0,y_0)$ and line that connect $A(x_1,y_1)$ and $B(x_2,y_2)$ calculated in equation 10.

(10) $$d = \frac{|(x_2-x_1)(y_1-y_0)-(x_1-x_0)(y_2-y_1)|}{\sqrt{(x_2-x_1)^2-(y_2-y_1)^2}}$$

As shown in figure 6, convex points that are after or before convex defect are eligible to be fingertips. We call this set, EP. We find average of convex defects to COG and call it CDAvg. Points of EP that their distance are more than CDAvg + 25*CDAvg/100, they assumed as fingertips.

If the number of fingertips are more than one, gesture is not pointing.

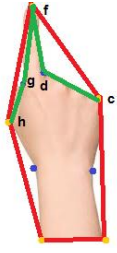

Figure 6. Pointing gesture

Figure 6 shows a pointing gesture. If there is one fingertip, which satisfy below conditions, we can conclude this fingertip is index fingertip.

- $80 \leq \widehat{fdc} \leq 130$
- $190 \leq \widehat{fgh} \leq 170$

### B. Finger Orientation Calculation

We have three options to calculate pointing vector
- Using COG
- Using next convex defect
- Using bisects of previous and next point of fingertip in convex defects

One of the simplest ways of reporting the orientation of the finger is computing angle of the vector created by COG and the fingertip, but it is not reliable in all postures and there is a possibility to have an error about 30 degrees[9].

Another approach with better results is using the next convex defects after fingertip. Green line in figure 7 shows pointing orientation.

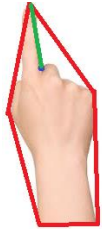

Figure 7. Pointing orientation using convex defects

For a better result, if $(x_f, y_f)$ is the fingertip, we report the bisect of angle created by three points, $(x_{f-1}, y_{f-1})$, $(x_f, y_f)$, and $(x_{f+1}, y_{f+1})$ as the finger orientation. Figure 8 shows the result of this approach.

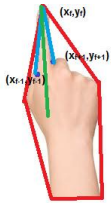

Figure 8. Finger Orientation Calculation

## III. EXPERIMENTAL RESULTS

In this section, we describe the performance of our hand pointing recognition procedure. The described procedure has been implemented and tested using Visual C++ and OpenCV libraries.

For the performance evaluation of the hand pointing recognition, the system has been tested on 50 different test sets , nd each test runs in different rooms with different light condition and context. In each experiment, we point to all 360 degrees for both pointing and non-pointing postures. Table 1 shows results of experiments with no light change.

Table 1. test results with no light change.

| False | True | |
|---|---|---|
| 2.85% | 95% | Positive |
| 4% | 91% | Negative |

Our results are more accurate in comparison to the results in [9]. Figure 9 and 10 shows some results in white and yellow light.

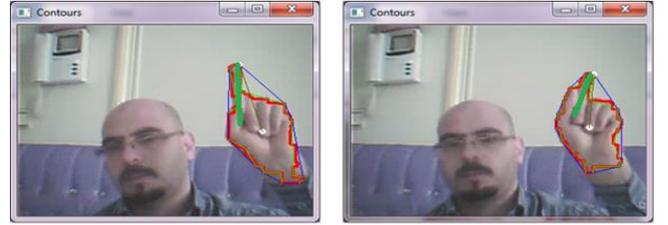

Figure 9. Result example in white light

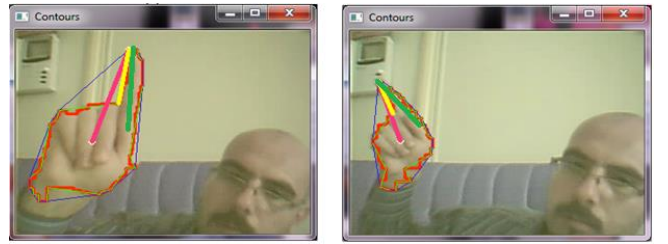

Figure 10. Results example in yellow light

We have also tested the system in light changing condition. Table 2 summarises test results with light change in a period of one minute.

Table 2. test results with light change.

| False | True | |
|---|---|---|
| 3% | 87% | Positive |
| 5% | 83% | Negative |

## IV. Conclusions

Hand posture and pointing recognition have numerous applications. In this research, we have proposed a vision-based hand pointing recognition system which recognizes hand pointing posture and also calculates the pointing angle in a plane perpendicular to the camera view direction.

Our approach doesn't force users to wear sensors or gloves, and it is applicable in a room with the variety of colors. This method consists of (1) Background Subtraction (2) Hand Segmentation using Histogram Template of Skin (3) Hand segmentation (4) Finger orientation calculation and (5) finger direction detection.

Next research will focus on more pointing classes and more complicated scenes.